\pdfoutput=1
\documentclass[letterpaper, 10pt, conference]{ieeeconf}
\IEEEoverridecommandlockouts
\usepackage{amsmath}
\usepackage{amssymb}
\usepackage{graphicx}
\usepackage{booktabs}
\usepackage{multirow}
\usepackage[dvipsnames]{xcolor}
\usepackage{caption}
\usepackage{cite}
\usepackage{hyperref}
\hypersetup{
  colorlinks=true,
  linkcolor=MidnightBlue,
  citecolor=BrickRed,
  urlcolor=RoyalBlue
}
\usepackage{cleveref}

\graphicspath{{figures/}}

\newcommand{\myparagraph}[1]{\vspace{0.02cm}\noindent\textbf{#1}}

\title{\LARGE \bf
The Neverwhere Visual Parkour Benchmark Suite
}

\author{%
  Ziyu Chen$^{1,2}$ \quad Henghui Bao$^{*1}$ \quad Haoran Chang$^{*3,6}$ \quad Alan Yu$^{4,5}$ \quad Ran Choi$^{4}$ \\
  Kai McClennen$^{4,5}$ \quad Gio Huh$^{5,7}$ \quad Kevin Yang$^{5,8}$ \quad Ri-Zhao Qiu$^{3}$ \quad Yajvan Ravan$^{4,5}$ \\
  John J. Leonard$^{4}$ \quad Xiaolong Wang$^{3}$ \quad Phillip Isola$^{4}$ \quad Ge Yang$^{\dagger 4,5}$ \quad Yue Wang$^{\dagger 1}$%
  \thanks{$^{*}$Equal contribution. $^{\dagger}$Equal advising. Work done while Ziyu Chen was a visiting student at USC before joining Stanford. $^{1}$University of Southern California, $^{2}$Stanford University, $^{3}$UC San Diego, $^{4}$Massachusetts Institute of Technology, $^{5}$FortyFive Labs, $^{6}$UC Los Angeles, $^{7}$California Institute of Technology, $^{8}$Harvard University.}%
}

\begin{document}
\maketitle
\thispagestyle{empty}
\pagestyle{empty}

\begin{abstract}
State-of-the-art visual locomotion controllers are increasingly capable at handling complex visual environments, making evaluating their real-world performance before deployment increasingly difficult. This work intends to narrow this train/evaluation gap by developing a collection of hyper-photo-realistic, closed-loop evaluation environments -- The Neverwhere Benchmark Suite -- comprised of over sixty 3D Gaussian Splatting reconstructions of urban indoor and outdoor scenes. Our goal is to encourage large-scale and reproducible robot evaluation by making it easier to create and integrate Gaussian splats-based reconstructions into simulated continuous testing setups. We also underscore the potential pitfalls of relying exclusively on 3D Gaussian-generated data for training, by providing policy checkpoints trained over multiple Neverwhere scenes and their performance when evaluated in novel scenes. Our analysis illustrates the necessity of sourcing diverse data to ensure performance. Code and data are available on the project page: \url{https://ziyc.github.io/neverwhere-bench/}.
\end{abstract}

\section{Introduction}
The past few years witnessed a rapid acceleration in progress in robotics. Data-driven, general-purpose learning algorithms, which treat specific tasks as data points from a general problem class, proved to be the scalable approach to producing robots that are robust, capable, and intelligent. As our robots graduate the confined lab environments to face the open world, real-world evaluation and hand-crafted simulation environments are proving insufficient. We need an evaluation strategy that is equally scalable to quantify progress. How do we create abundantly diverse and realistic environments to test our robot?

This work aims to develop a scalable approach to testing real-world visuomotor policies in automated, closed-loop simulations. We focus on visual locomotion in legged robots as our test bed, a class of robotic tasks where perception is tightly coupled with actions. We start with a domain where the 3D environment is complex but the physics is relatively simple. Our main contribution is Neverwhere, a collection of over 60 high-fidelity digitally recreated scenes that covers diverse urban structure, including stairs, speed bumps, indoor carpeted lab spaces and the outdoor, with and without vegetation. An equally essential objective is to empower the community to build their own set of benchmarks.

\begin{figure}[t]
\includegraphics[width=\linewidth]{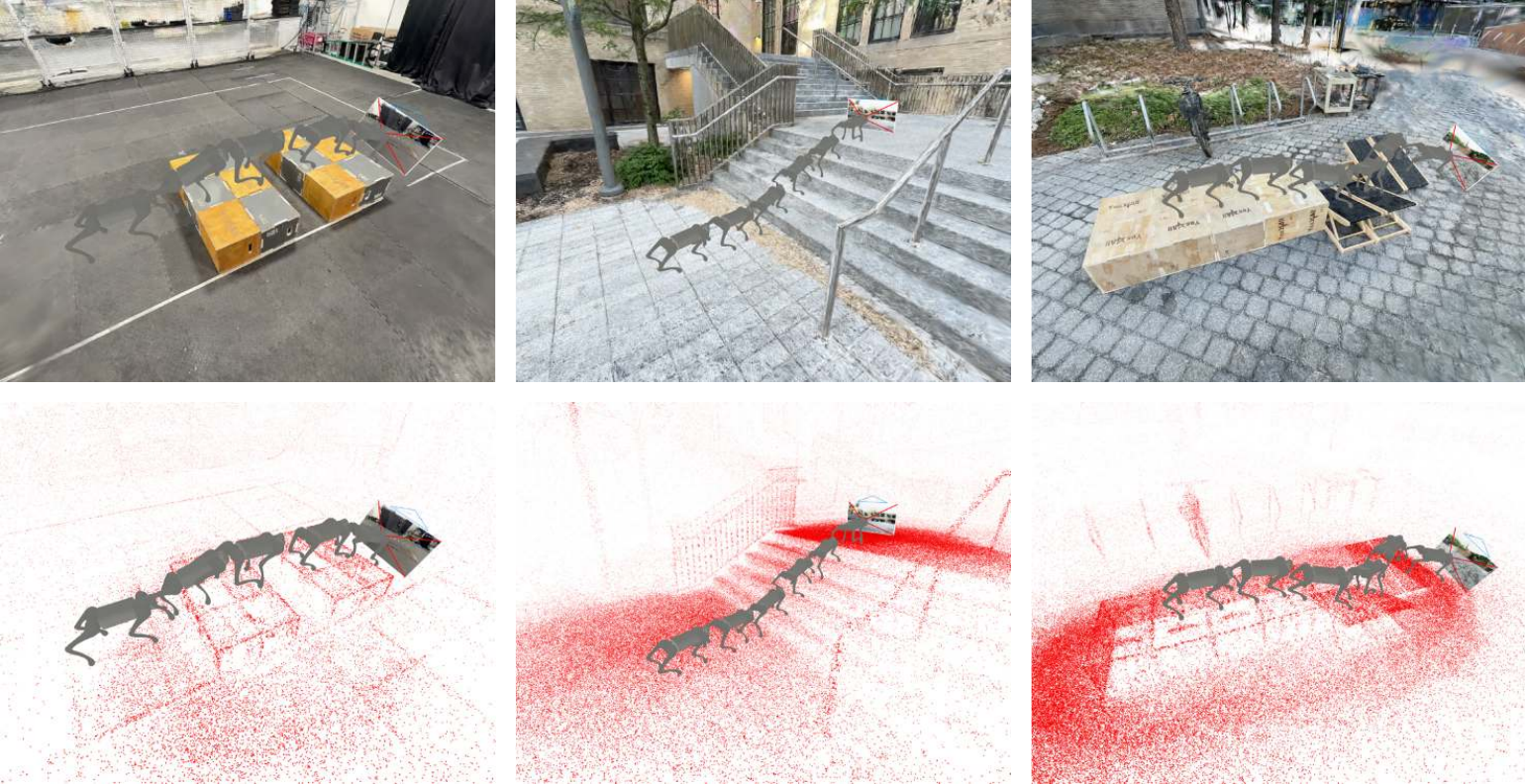}
\caption{\textbf{The Neverwhere Benchmark Suite.}
  We offer over \textit{sixty} high-quality Gaussians-based physical environments, and the Neverwhere graphics toolchain for producing accurate scene collision mesh and visual representations. Our aim is to promote reproducible robotics evaluation via fully automated, continuous testing in closed-loop environments. RGB Images are rendered from 3D Gaussians.}
\label{fig:gallery}
\end{figure}

The Neverwhere toolchain addresses three essential challenges in building evaluation environments for robots: The primary challenge is to capture the world in its full messiness which exceeds the expressivity of traditional 3D mesh. The second challenge is that in practice, quadruped robots observe the world from an angle that sits out of the distribution of human camera views. This coupled with the extraordinary expressivity of the Gaussian substrate, results in poorly rendered ego camera input. The final challenge is about geometry. It remains difficult, in practice, to obtain detailed collision mesh from 3D Gaussian that are modeled using hand-held mobile videos. We addressed them by developing a better framework with improved geometry for 3D Gaussians and fine-grained collision mesh that takes advantage of traditional multi-view stereo reconstruction.
Our contributions are summarized as follows:
\begin{itemize}
\item \noindent We introduce the Neverwhere benchmark suite, featuring over 60 high-quality environments powered by 3D Gaussians, encompassing a diverse range of urban indoor and outdoor scenes.
\item \noindent We present a data collection toolchain that facilitates the generation of new benchmark environments with minimal human intervention, allowing users to create reconstructed scenes directly from uncalibrated captures.
\item \noindent We provide baseline visual policy checkpoints trained on Neverwhere 3D Gaussian environments to facilitate future research on both the limitations and the potential of simulation-based training for real-world transfer. 
\end{itemize}

\begin{figure*}[t]
\centering
\includegraphics[width=\textwidth]{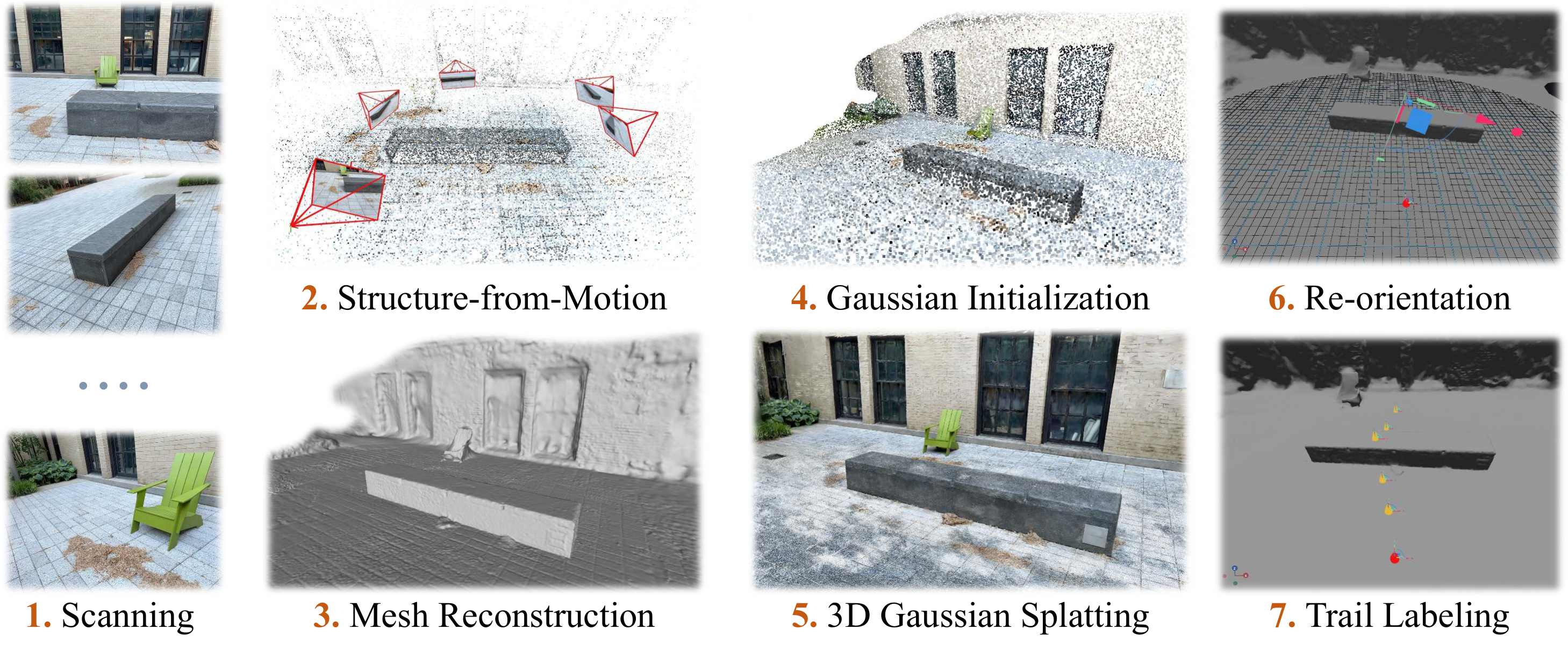}

\caption{\textbf{Neverwhere Toolchain.} The toolchain takes \textcolor{brown}{(1)} multi-view images as input and follows a sequential process: 
\textcolor{brown}{(2)} A Structure-from-Motion modules is applied to obtain camera calibrations, 
\textcolor{brown}{(3)} An optimization-based MVS module is used to estimate the scene geometry, 
\textcolor{brown}{(4)} Points are sampled from the textured meshes, which are then used as initializations for \textcolor{brown}{(5)} training 3D Gaussians to model the scene. 
Once the scene modeling is complete, human input is required for 
\textcolor{brown}{(6)} reorienting the mesh to align with scene conventions and 
\textcolor{brown}{(7)} labeling waypoints for visual parkour policies.}
\label{fig:toolchain}
\end{figure*}

\section{Related Work}

Robotics has long relied on physics simulators for policy evaluation~\cite{todorov2012mujoco}. 
Recent learning-based approaches increasingly depend on high-fidelity visual inputs, raising the bar for simulation realism~\cite{chi2023diffusion}. Both the quality and the range of physics and materials have improved significantly in more recent physics engines such as IssacSim~\cite{IsaacSim} and ManiSkill3~\cite{gu2023maniskill2,taomaniskill3}, extending coverage to deformable material, fluid, and caustics. Despite these improvements, lack of 3D content remains a bottleneck. Recent benchmark efforts  significantly raised the bar: BiGym~\cite{chernyadev2025bigym}, for instance, provided a high-quality, manually CAD'ed 3D collision mesh for an articulated dish washer. At the scene level, RoboCasa~\cite{robocasa2024} provides fourteen manually designed kitchen scenes.

Neverwhere differs from these prior efforts~\cite{nasiriany2024robocasa,li2023behavior,james2019rlbench,gupta2023robohive,chernyadev2025bigym}
in two ways: First, advancements in neural scene representation made investment in traditional assets and lighting setup less critical. Neverwhere uses 3D Gaussian Splatting to replace mesh-based rendering, which not only simplifies construction but also enables the creation of highly detailed digital replicas, which can be done by the end-user using Neverwhere's open-source toolkit.
Second, Neverwhere builds upon the MuJoCo~\cite{todorov2012mujoco} physics engine and aims to provide accurate collision geometry. It does so without requiring LiDAR sensors and depth measurements. The closest are autonomous driving simulators scenes that are used for closed-loop evaluation of autonomous vehicles~\cite{yang2023unisim,chen2024omnire}. Recently, concurrent works such as VR-Robo~\cite{zhu2025vr}, GaussGym~\cite{escontrela2025gaussgym}, and GSworld~\cite{jiang2025gsworld} also explored neural representations for constructing real-to-sim environments across various embodiments including quadrupeds, humanoids, and manipulation. Developed in parallel with them, Neverwhere provides an open-source foundation for building customizable evaluation setups in robotics, with plans to incorporate more advanced capabilities in the future.

Generative AI is increasingly being used in robotics to generate task scenarios, rewards, and assets. Platforms such as RoboCasa~\cite{robocasa2024}, MimicGen~\cite{Mandlekar2023mimicgen} and video world models such as Cosmos~\cite{kim2025cosmospolicy} and Veo~\cite{veorobotics2025} help increase the diversity of training data and environments. However, Neverwhere takes a different approach by emphasizing photo-realism and accurate modeling over diversity. We argue that, when it comes to evaluation benchmarks, a deterministic and curated set of challenging test cases prioritizes signal over noise. Neverwhere's use of 3D Gaussian splats is better suited for evaluation than for training, as the former favors reduced variability.

\section{From Photons to Physical Twins}\label{sec:pipeline}

In this section, we introduce the Neverwhere toolchain for scene creation. We aim to develop a scalable and efficient framework to enable users to easily generate their own digital twins. While existing 3DGS techniques~\cite{3dgs, mipsplat, mcmc_splat, sugar, 3dgut, gaussian_ray_tracing, 2dgs, scaffoldgs} excel at visual fidelity, but the resulting meshes may lack the physical accuracy. Robots require precise collision geometry to be evaluated correctly. Developing high-fidelity digital replicas for robot simulation necessitates both minimal visual gap and accurate scene geometry modeling for reliable physics-based interactions. 

Beyond visual appearance and geometry, we also emphasize system robustness to simplify scene creation and lower the barrier for generating digital twins, even under imperfect capture conditions (e.g., poor lighting or shadow areas). To this end, we propose a unified yet highly modular framework that integrates recent advances in scene calibration, reconstruction, and modeling, along with a suite of built-in tools to address scalability issues across diverse scenarios. Our framework achieves automatic processing that converts uncalibrated multi-view images into a physical digital scene with geometrical-improved 3D Gaussians along with spatially aligned high-quality collision geometry. The resulting twins are correctly oriented and scaled to metric size for deployable robot simulation. The overall pipeline and process is shown in \cref{fig:toolchain}, we describe each component of our toolchain as follows:

\myparagraph{Formulation.} Given a set of $N$ uncalibrated images $\mathcal{I} = \{ \mathbf{I}_i\}_{i=1}^N$, a camera pose module $\Theta$ is used to estimate their poses (\cref{fig:toolchain}-(2)), yielding $\mathcal{P} = \{ \mathbf{P}_i\}_{i=1}^N$. Subsequently, a mesh reconstruction module $\Phi(\mathcal{I}, \mathcal{P})$ is employed to recover the scene geometry $\mathcal{M}$ from calibrated images(\cref{fig:toolchain}-(3)).
Following this, a Gaussian Splat module $\Psi$ is applied for scene appearance modeling (\cref{fig:toolchain}-(5)). However, vanilla Gaussian splatting tends to overfit the training views, often producing suboptimal novel-view renderings when the viewpoint deviates significantly from the input trajectories (e.g., views from a quadruped robot versus typical handheld captures). Introducing additional geometric supervision such as depth from auxiliary sensors~\cite{ren2024agsmesh,turkulainen2024dnsplatter} can help but increases system complexity. To achieve comparable regularization without extra sensors, we leverage depth extracted from intermediate representations of $\Phi$ to guide the Gaussian optimization. This yields a geometry-grounded Gaussian trainer $\Psi(\mathcal{I}, \mathcal{P}, \mathcal{M})$, formulated with the following loss:
\begin{equation*}
    \mathcal{L} = \left(1-\lambda_r\right) \left\|\mathbf{I}-\hat{\mathbf{I}}\right\|_1+\lambda_r \mathcal{L} _{\text {SSIM}} + \lambda_D \left\|\mathbf{C}\odot(\mathbf{D}-\hat{\mathbf{D}})\right\|_1
\end{equation*}
Here, $\mathbf{D}$ and $\mathbf{C}$ denote the depth and confidence maps extracted from the patch-matched geometric cache of $\Phi$. Thanks to our modular framework, we can continuously leverage the strengths of emerging state-of-the-art techniques. In the current implementation, we employ COLMAP~\cite{schoenberger2016sfm,schoenberger2016mvs} as $\Theta$ for camera calibration and OpenMVS~\cite{openmvs2020} as $\Phi$ for mesh reconstruction. 

\myparagraph{Scene alignment and task annotation.}
Since $\Theta$ estimates camera poses up to an arbitrary scale and orientation, the reconstructed splats and meshes are misaligned in scale, orientation, and coordinate convention (z-up), making them unsuitable for direct robot simulation. We introduce an intuitive labeling tool that allows users to quickly rescale and reorient the scene to real-world coordinates (\cref{fig:toolchain}-(6)) and define task annotations such as waypoint trails for policy evaluation (\cref{fig:toolchain}-(7)). The entire process is highly efficient, taking less than two minutes (see the project page for demos).

\begin{figure}[t]
\centering
\includegraphics[width=\linewidth]{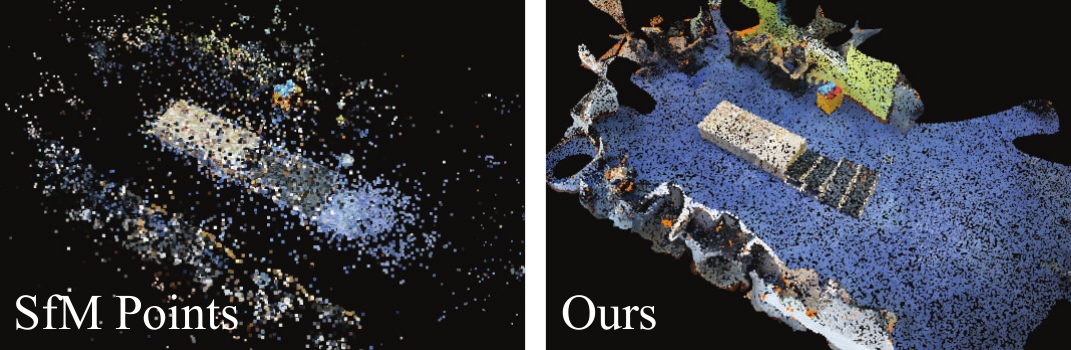}
\caption{\textbf{Gaussian Initialization.} SfM points are sparse and noisy, often causing artifacts in free space, while points sampled from textured meshes are well distributed and surface-aligned, yielding improved geometric accuracy of the Gaussians.}
\label{fig:comp_gs_init}

\vspace{0.8em}
\includegraphics[width=\linewidth]{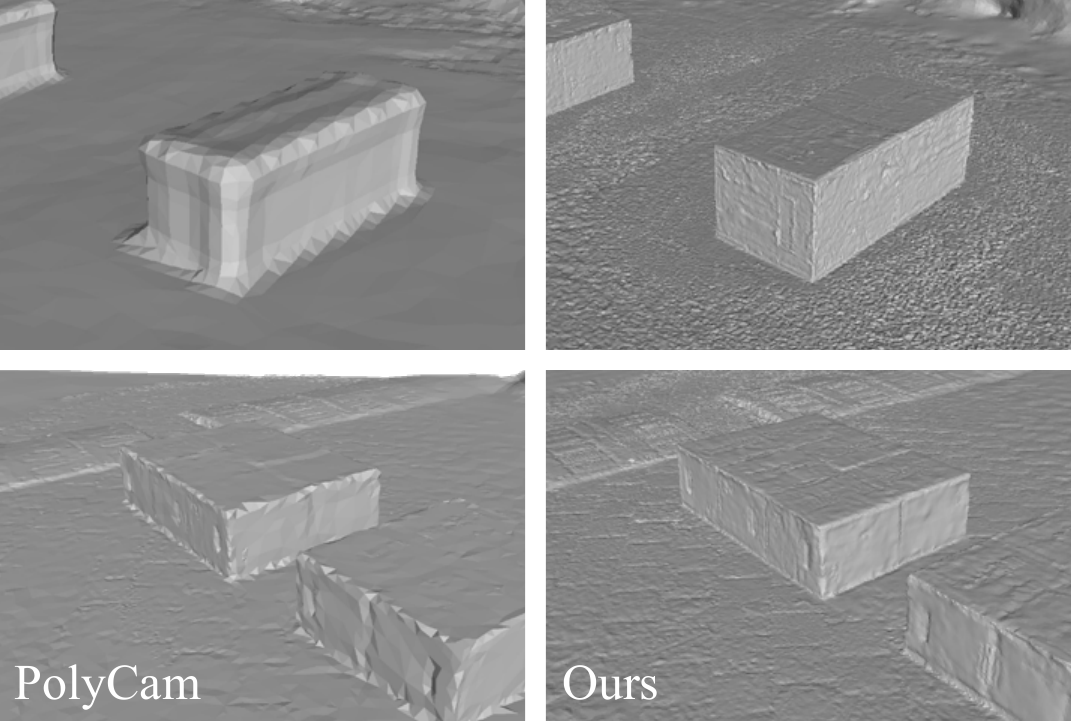}
\caption{\textbf{Comparison on Scene Geometry.} Left: mesh captured by the consumer 3D scanning app PolyCam. Right: reconstruction produced by our pipeline.}
\label{fig:comp_mesh}

\vspace{0.3em}
\captionsetup{type=table}
\caption{\textbf{Quantitative comparison of integrated 3DGS models.} Scene quality of \textit{Splatfacto} and \textit{Splatfacto-MCMC} integrated within our framework, showing that our training strategies benefit the integrated scene modeling methods. P/S/L denote PSNR, SSIM, and LPIPS, respectively.}
\label{tab:splatfacto}

\setlength{\tabcolsep}{2.8pt}
\renewcommand{\arraystretch}{1.1}
\begin{tabular}{l|ccc|ccc}
\toprule
\multirow{2}{*}{Models} &
\multicolumn{3}{c|}{\textbf{Reconstruction}} &
\multicolumn{3}{c}{\textbf{Novel View}} \\
 & P $\uparrow$ & S $\uparrow$ & L $\downarrow$ & P $\uparrow$ & S $\uparrow$ & L $\downarrow$ \\
\midrule
Splatfacto-Orig. & 32.30 & 0.919 & 0.136 & 30.21 & 0.886 & 0.172 \\
Splatfacto-Ours  & \textbf{32.52} & \textbf{0.925} & \textbf{0.122} & \textbf{30.32} & \textbf{0.891} & \textbf{0.161} \\
\midrule
mcmc-Orig.       & 33.38 & 0.944 & 0.094 & \textbf{31.02} & 0.915 & 0.127 \\
mcmc-Ours        & \textbf{33.45} & \textbf{0.946} & \textbf{0.089} & 30.92 & \textbf{0.915} & \textbf{0.124} \\
\bottomrule
\end{tabular}

\end{figure}

\myparagraph{Geometry-guided Gaussian initialization.}
To achieve higher-quality 3DGS with improved multi-view consistency and surface alignment, we initialize the Gaussians using colored point samples (\cref{fig:toolchain}-(4)) from the collision mesh $\mathcal{M}$, which provides stronger geometric priors. This initialization is more structurally coherent than using scattered SfM points alone, as illustrated in \cref{fig:comp_gs_init}, resulting in geometry-consistent renderings from robot viewpoints.

\myparagraph{Enhanced geometry from mobile captures.}
Accurate collision geometry is critical for precise contact and physics simulation. A common approach is to employ on-device multi-view stereo tools. However, geometry produced by mobile applications (e.g., Polycam~\cite{polycam}) often lacks the fine details required for high-fidelity contact simulation due to the limited computational capacity of mobile devices. Moreover, many mobile reconstruction methods rely on depth sensors for robustness. As shown in \cref{fig:comp_mesh}, meshes reconstructed with OpenMVS~\cite{openmvs2020} exhibit finer geometric details and more complete scene coverage, surpassing those generated by on-device mobile software. This also conveniently provides the depth supervision needed to improve our Gaussian representation, without any additional cost or sensing effort, as discussed above.

\myparagraph{Experiments on assets quality.}
To evaluate the quality of reconstructed scenes from our end-to-end pipeline and to demonstrate the flexibility of our modular scene-creation framework, we integrated Nerfstudio~\cite{nerfstudio} as the 3DGS module within our system. In this configuration, Nerfstudio benefits from our intermediate representations produced by the mesh reconstruction module $\Phi$, enabling geometry-guided regularization and mesh-based Gaussian initialization.
We tested on two tasks: Scene Reconstruction (trained and evaluated on all images) and Novel View Synthesis (trained on 90\%, evaluated on 10\%). The average results for a batch of 7 scenes are reported in \cref{tab:splatfacto}, where \textit{Ours} indicates applying both OpenMVS-based initialization and depth supervision. The quantitative results show that our MVS-first, geometry-guided framework consistently boosts the performance of the integrated 3DGS models. 

\myparagraph{Implementation.} In practice, we integrated \textit{Splatfacto-MCMC} as our 3DGS module. Following the Neverwhere scene creation pipeline described above, we scanned and processed over \textit{sixty} high-quality, physically-aware 3D Gaussian environments for closed-loop visual parkour evaluation.

\section{The Neverwhere Benchmark Suite}

Neverwhere provides over \textit{sixty} parkour scenes for quadruped. These physical environments are created from 3D scans of two university campuses, covering both outdoor and indoor domains with diverse obstacle layouts and visual appearances. The tasks were written in Python, and physics simulation is implemented with MuJoCo \cite{todorov2012mujoco}. The locomotion setup follows:

\myparagraph{Action Space.} The action space consists of twelve target joint positions for the quadruped robot, with each of the robot's four legs having three actuated joints: hip abduction/adduction, hip flexion/extension, and knee flexion/extension.

\myparagraph{Observation Space.} The observation includes the robot's ego state, \(\mathbf{e} = \{v, \mathbf{q}, \dot{\mathbf{q}}\}\), where \(v\) is the linear velocity, \(\mathbf{q}\) the joint position, and \(\dot{\mathbf{q}}\) the joint velocities.
For evaluating visual policies, we provide visual observations in different data modalities, including RGB renders from gsplat, depth maps, point clouds, and semantic maps. The rendering pipeline detailed in \cref{sec:rendering_wrappers} provides these diverse data modalities for visual policy input.
We provide privileged observation for training purpose, including a heightmap of the scene, offering a top-down view of the terrain, which can be further processed into ScanDots for lightweight inference~\cite{cheng2023parkour}. The moving direction, represented as a single angle value, can be used to guide navigation.

\subsection{Evaluation Tasks}
We design four scenarios to evaluate locomotion generalization across increasing difficulty~\cite{cheng2023parkour}: 
(1) \textbf{Hurdles}: traversing fixed-height obstacles; 
(2) \textbf{Gaps}: jumping across openings; 
(3) \textbf{Ramps}: walking on inclined surfaces; 
(4) \textbf{Stairs}: ascending staircases.

\begin{figure}[t]
\includegraphics[width=\linewidth]{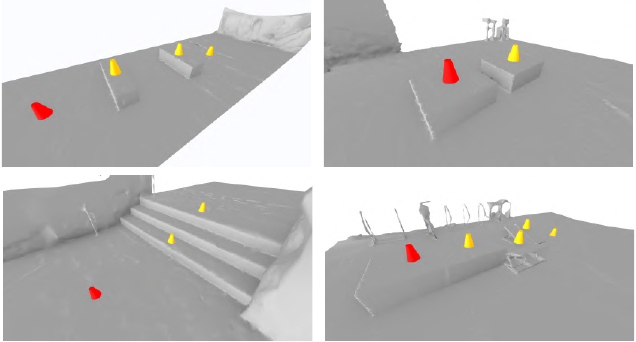}
\caption{\textbf{Annotated Trails.} Way-points are annotated for each scene to define each evaluation task. \textbf{\color{yellow}Yellow cones} indicate waypoints, whereas \textbf{\color{red}{red} cones} are the starting points.}
\label{fig:collision_geometry}
\end{figure}

\myparagraph{Evaluation Metrics.}
Each scene is annotated with waypoints defining a reference trail. Performance is measured by \textit{Success Rate}, defined as the percentage of waypoints reached. See~\cref{fig:collision_geometry} for examples.

\begin{figure}[t]
\centering
\includegraphics[width=\linewidth]{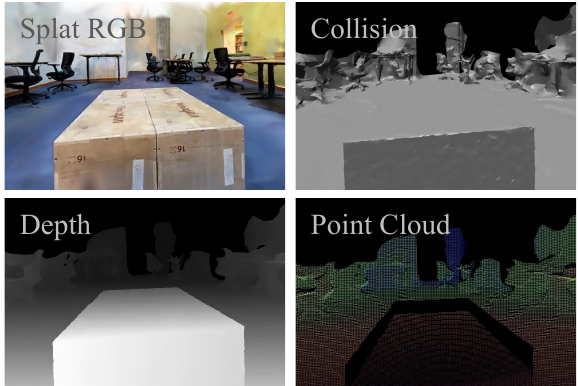}
\caption{\textbf{Rendering Wrappers.} We designed a rendering pipeline that provides diverse wrappers for multi-modal observations, including but not limited to: color images from Gaussian splats, depth maps, heightmaps, and LiDAR projections, to support a wide range of visual based policies.}
\label{fig:wrappers}
\end{figure}

\begin{figure*}[t]
\centering
\includegraphics[width=\linewidth]{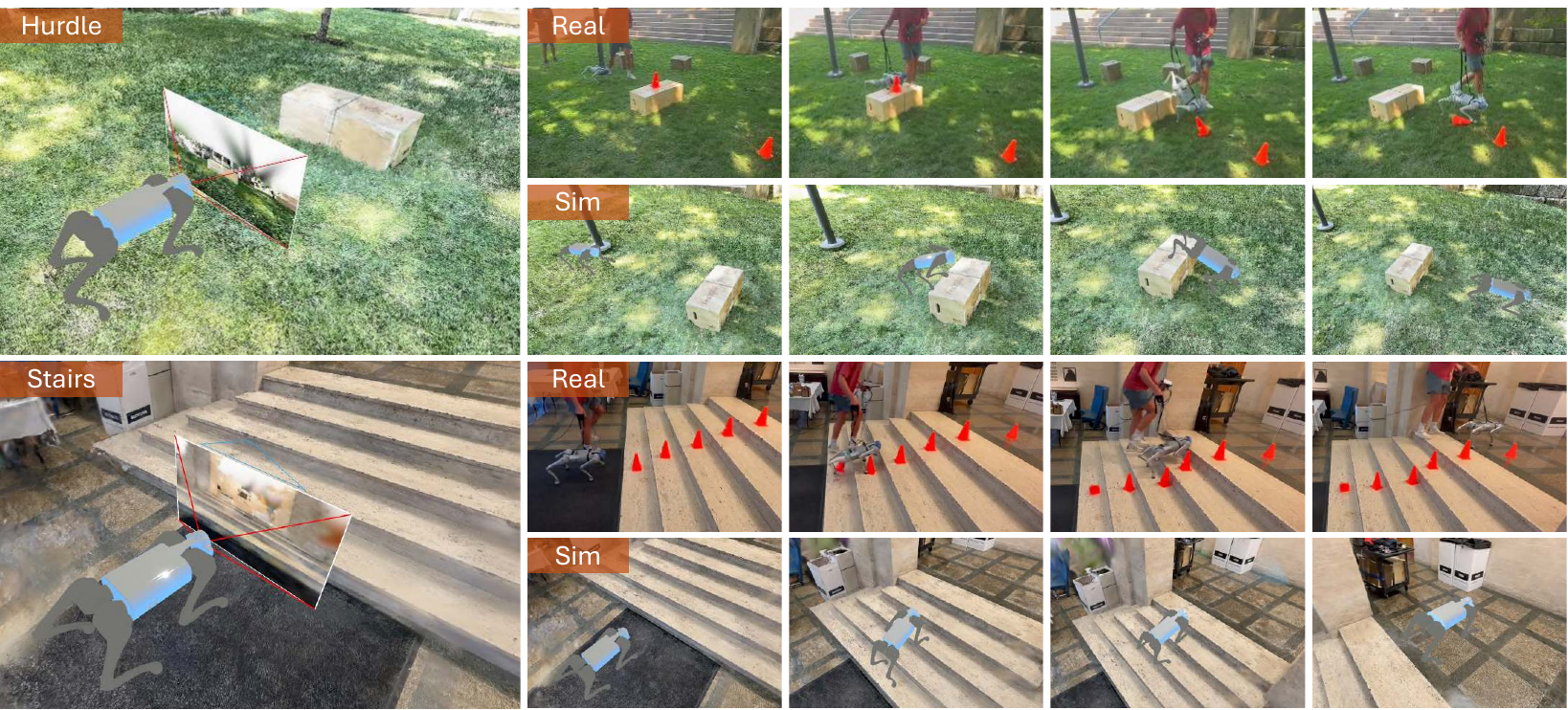}
\caption{\textbf{Visualization of sim-to-real parallel evaluation.} By examining a visual policy's behavior and its action/failure patterns in the Neverwhere simulation environments, we can obtain early indications of its real-world performance. }
\label{fig:real-sim}
\end{figure*}

\myparagraph{Scene Construction.}
Hurdle scenes are created using a sequence of low barriers that the robot must traverse, with waypoints placed atop each obstacle to define the intended path. Some scenes additionally leverage natural outdoor structures (e.g., stone benches) as hurdles. Gap scenes consist of two elevated platforms separated by a fixed opening, requiring the robot to perform jumping behaviors. Waypoints are placed at the center of each platform. Ramp scenes include multiple inclined boards arranged in staggered layouts, testing locomotion stability on sloped terrain. Stair scenes use real staircases of varying height, material, and texture, captured in both indoor and outdoor environments, with waypoints annotated along the ascent.

\begin{table}[t]
\centering
\caption{Parallel evaluation of \textit{LucidSim}~\cite{yu2024lucidsim}, with 10 trials per scene and reported completion rates.}
\label{tab:eval_lucidsim}
\resizebox{0.35\textwidth}{!}{
\begin{tabular}{lccc}
\toprule
\textbf{Tasks} & \textbf{Trials} & \textbf{Real-world} & \textbf{Neverwhere} \\
\midrule
Hurdles & 20 & 75\% & 75\% \\
Stairs  & 20 & 60\% & 85\% \\
\bottomrule
\end{tabular}
}
\end{table}

\subsection{Rendering Wrappers}\label{sec:rendering_wrappers}
Visual-locomotion policies perform well with depth inputs~\cite{cheng2023parkour,agarwal2023legged,luo2024pie} but are still limited with RGB observations. To advance the development of robust RGB-based policies, a scalable evaluation testbed with diverse visual modalities is needed. Recent works that use generative models to generate RGB images for visual policy training ~\cite{yu2024lucidsim} have shown impressive real-world results, in part enabled by 3DGS-based simulation for evaluation. Neverwhere aims to significantly extend the scale and diversity of such evaluation setups and provides a unified set of rendering wrappers that support multiple visual modalities:

\myparagraph{Gaussian Splatting.} We adopt 3DGS~\cite{kerbl3Dgaussians} for photorealistic rendering to reduce the domain gap in policy training and evaluation. 3DGS provides high-quality observations and real-time performance, making it suitable for closed-loop evaluation. To incorporate task-relevant targets (e.g., cones) absent from the original 3DGS scene, we blend them into the rendered view using semantic masks from MuJoCo~\cite{todorov2012mujoco}. Our experiments (in~\cref{sec:exp_multi_scene}) show that these visual cones provide strong cues that lead to better policies.

\myparagraph{Depth Map.}
Depth is obtained by converting MuJoCo-rendered depth maps~\cite{todorov2012mujoco} into MiDaS-style inverted depth~\cite{Ranftl2022}. These depth maps serve as policy observations and can also condition depth-guided generative models~\cite{zhang2023adding}, enabling robust training and zero-shot transfer to real-world RGB inputs~\cite{yu2024lucidsim}.

\noindent Beyond these, other modalities including semantic masks, heightmaps, and LiDAR point clouds are supported, as shown in~\cref{fig:wrappers}. The system also allows users to add their own wrappers, for instance, a depth-conditioned visual generation wrapper for enhanced visual diversity.

\section{Sim-to-Real Consistency of Neverwhere}

We performed parallel real-world and simulation evaluations to assess whether Neverwhere can reliably indicate a policy's real-world performance (Fig.~\ref{fig:real-sim}). Using LucidSim~\cite{yu2024lucidsim} as the baseline policy, we constructed four scenes for the tasks it supports (hurdles and stairs). For each scene, we built a corresponding 1:1 Neverwhere simulation environment and rolled out the same policy in both settings. The results are summarized in Tab.~\ref{tab:eval_lucidsim}.
For the simpler hurdles task, the policy achieves comparable performance across both environments, suggesting that the visual gap is small in our benchmark. For the more challenging stairs task, Neverwhere reports higher success rates than real-world, likely due to increased physical complexities present in real-world stairs. Overall, the aligned performance trends indicate that Neverwhere provides a meaningful proxy for evaluating visual-policy behavior.

\begin{figure*}[t]
\centering
\includegraphics[width=\linewidth]{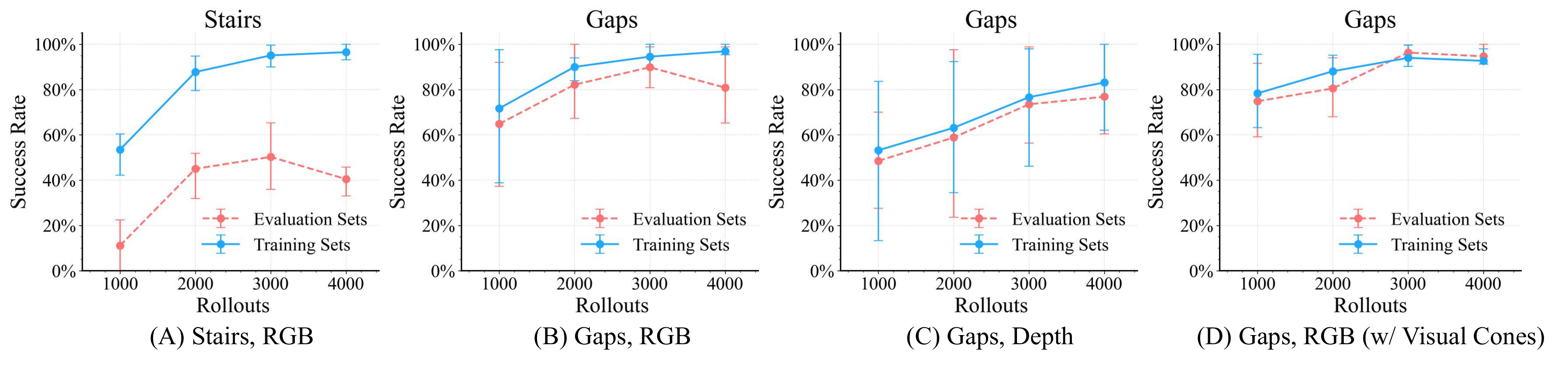}

\caption{\textbf{Results on Multi-Scene Closed-Loop Training.} We split each task's scenes into 70\% for training and 30\% for evaluation. For each scene, we perform 50 rollouts and report the average success rate over all rollouts in the train and evaluation sets. See the project page for more closed-loop training results.}
\label{fig:results_multi_scene}

\end{figure*}

\begin{figure}[t]
\centering
\includegraphics[width=\linewidth]{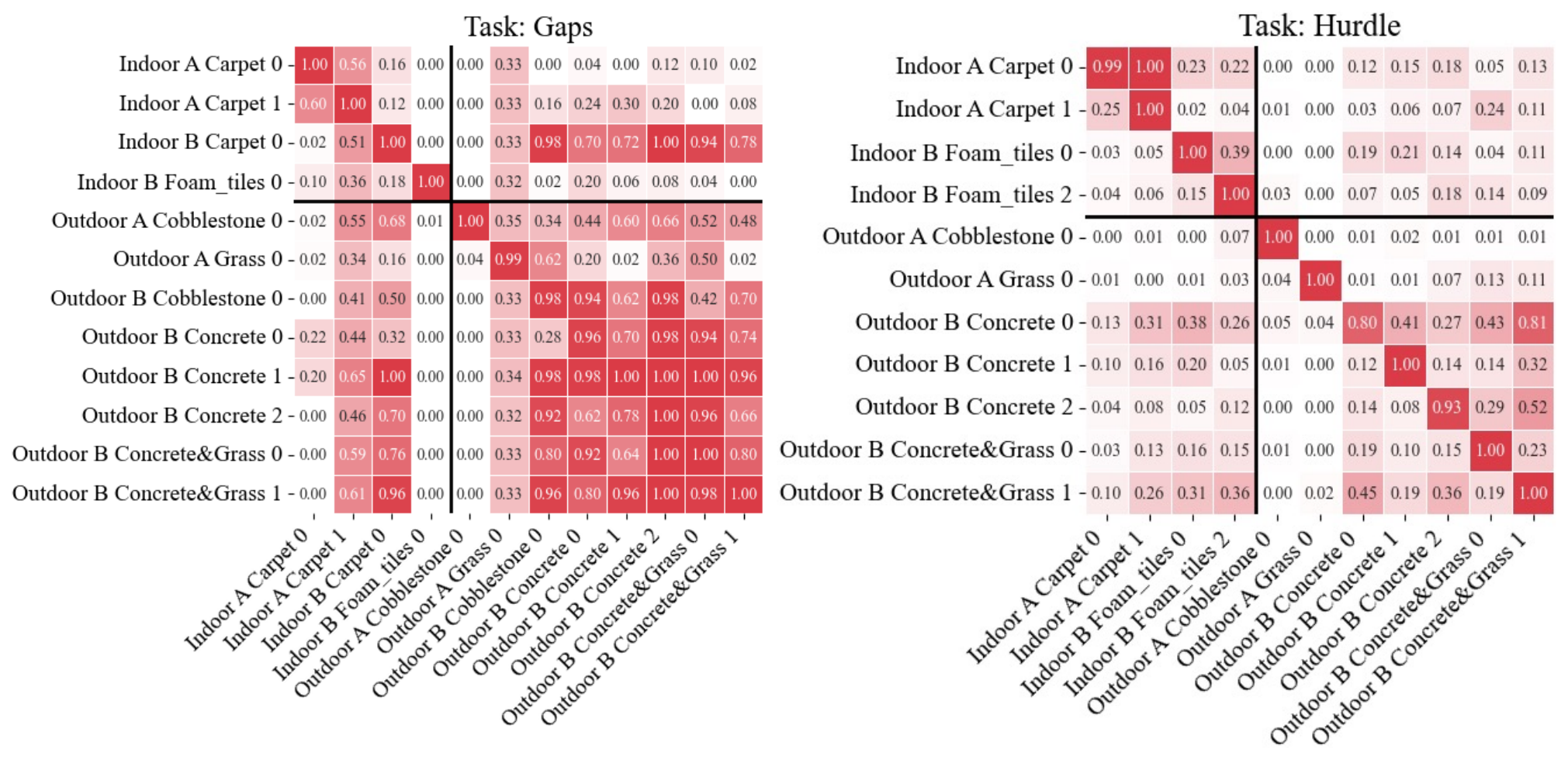}
\caption{\textbf{Performance Visualization of Single-Scene Policies.} Each row corresponds to a policy trained in a single scene and evaluated on all other scenes. Cross-scene evaluation (separated by the black line) shows clear clustering.}
\label{fig:results_single_scene}

\end{figure}

\section{Closed-loop training on Neverwhere}\label{sec:exp_multi_scene}

Our design intention is for Neverwhere to be used as part of an automated, continuous testing setup that quickly and scalably uses closed-loop simulation to assess the visual policy \textit{before} real-world deployment. Rather than serving as a large-scale training environment, Neverwhere is designed to function primarily as an evaluation benchmark, with scene diversity tailored for assessment rather than training. Training and testing environments have different requirements: the former benefits from system coverage and entropy, while the latter is best conducted deterministically to maximize interpretability. That said, in addition to evaluating policies on our benchmark, we also used Neverwhere for closed-loop training experiments to obtain a clear empirical answer.

\myparagraph{Training Setup.} We trained visual locomotion policies in our benchmark environments using a behavior-cloning setup guided by a privileged teacher, which we re-trained following \cite{cheng2023parkour}. We trained both depth- and RGB-based student policies, with RGB observations rendered through the Gaussian Splats wrapper. We adopt Action Chunking Transformers (ACT) \cite{zhao2023learning} as the student policy architecture to better handle challenging tasks such as gaps and ramps. We tried two settings:

\myparagraph{(1) Single-scene Training.} We trained each policy in a single environment using DAgger, collecting 1,000 trajectories per iteration and running four DAgger iterations in total. We evaluated the trained policy both on the training scene and on other scenes within the same task. Domain randomization was disabled for this setting.

\myparagraph{(2) Multi-scene Training.} To evaluate how well a policy generalizes across scenes within the same task, we conducted multi-scene training experiments. For each task, we split the scenes into 70\% for training and 30\% for evaluation. Domain randomization was applied during training: for depth policies, we injected random noise and randomly zeroed pixels; for RGB policies, we applied random rotations, cropping, Gaussian blur, and color transformations. All experiments were conducted on the Unitree Go1 robot.

\myparagraph{Single-Scene Training.}
We first tested whether a visual policy can fit well in a single domain.
We evaluated the policy on two tasks: Hurdles (easy) and Gaps (medium).
The policy is trained on one scene per task and evaluated on the remaining scenes.
To introduce variation, we add random noise to the trajectories so that evaluation rollouts differ from the training ones. As shown in \cref{fig:results_single_scene}, the policy performs well on the training scene but generalizes poorly on unseen scenes, which matches our expectation as the training domain is limited to a single scene. Notably, performance slightly improves on some unseen scenes that share similar visual characteristics (e.g., both being outdoor environments) with the training domain, as observed in the bottom-right corner of each confusion matrix in \cref{fig:results_single_scene}.

\myparagraph{Multi-Scene Training and Transfer.}
We further investigate how well visual policies generalize and transfer when trained across multiple scenes within the same task.
For the Stairs task (\cref{fig:results_multi_scene}-(A)), the gap between training and evaluation performance is large (about 50\% on average), whereas for the Gaps task (\cref{fig:results_multi_scene}-(B)) it is relatively small (about 10\% on average). This indicates that visual policies trained on our benchmark exhibit limited generalization, particularly for more challenging tasks.

\myparagraph{Ablation on Observation Types.}
We next compare two types of observations.
(1) RGB vs. Depth: As shown in \cref{fig:results_multi_scene}-(B) and (C), both inputs are trained with domain randomization. While RGB yields moderate results, Depth performs worse on the training set, indicating that depth-based policies are harder to train in this closed-loop setting.
(2) With vs. without visual cones: In the previous experiments, visual observations did not include cones, as none were placed in the scenes. Comparing \cref{fig:results_multi_scene}-(B) and (D), adding visual cones substantially improves training efficiency and overall performance on both training and evaluation sets. This shows that explicit visual cues (e.g., cones) help policies learn more robustly across scenes.

\section{Conclusion} 
\label{sec:conclusion}
We introduced Neverwhere, a benchmark suite and real-to-sim toolchain designed to provide a scalable and reliable testbed for evaluating visual policies before real-world deployment. As robot policies become increasingly capable while existing evaluation techniques lag behind, this work offers a complementary foundation for scalable and practical robot evaluation.

Although developed primarily for locomotion, the underlying toolchain, which is capable of producing contact-aware digital twins and multimodal observations from real environments, extends naturally to broader robotics domains. Neverwhere provides a unified and extensible foundation that supports real-to-sim-to-real research at scale.

\myparagraph{Limitations.} Neverwhere currently provides over 60 diverse scenes, and while much of the pipeline is automated, adding new environments still requires some manual effort. Reconstructed 3D splats and meshes are not automatically aligned to real-world scale or to a standard z-up convention, so limited manual reorientation, rescaling, and task labeling are needed before the scenes can be used in simulation. Future work will explore learning-based methods for automatic alignment and scene labeling of 3D reconstructions.

\section*{ACKNOWLEDGMENT}

This work was partially supported by the National Science Foundation through NSF CPS \#2434460. The USC Physical Superintelligence Lab acknowledges generous support from Toyota Research Institute, Dolby, Google DeepMind, Capital One, Nvidia, and Qualcomm. Yue Wang is also supported by a Powell Research Award.

\bibliographystyle{IEEEtran}
\bibliography{main}

\newpage
\makeatletter\setlength{\@fptop}{0pt}\makeatother

\appendices

\section{Scene Labeling Workflow}
\begin{figure}[h]
\centering
\includegraphics[width=\linewidth]{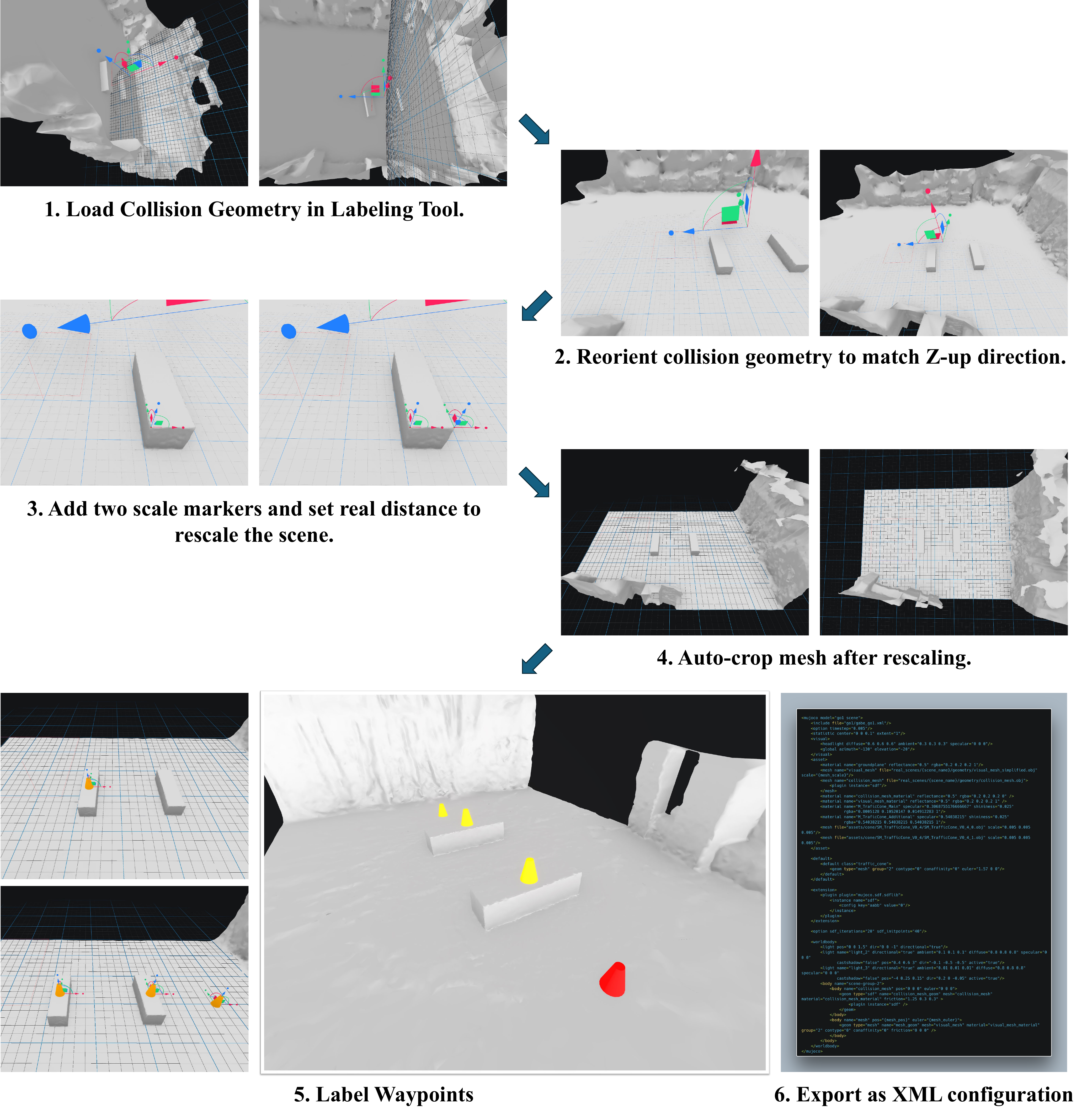}
\caption{\textbf{Overview of Scene Labeling Workflow.}}
\label{fig:supp_label}
\end{figure}

The Neverwhere toolchain automates the creation of high-fidelity physical digital twins from uncalibrated images and videos. However, due to random pose initialization, the resulting scenes often have arbitrary orientation and scale. Since robotics applications require a consistent frame of reference, especially to define gravity, some manual labeling is necessary. Thus, we designed an efficient labeling tool by integrating our annotation system with the visualization platform Vuer~\cite{vuer}. This streamlined design allows annotators to label a scene in approximately \textbf{one minute}, significantly improving the overall workflow efficiency. The full process, illustrated in \cref{fig:supp_label}, consists of the following steps:

\begin{enumerate}
\item Load the unprocessed collision geometry into the labeling system.
\item Manually rotate the geometry to align with the Z-up orientation ($\sim$20 seconds).
\item Place two markers on the mesh and input the real-world distance between them; the system automatically computes and applies the scale factor ($\sim$10 seconds).
\item The mesh is automatically cropped, no human input required.
\item Define waypoints for specific locomotion tasks ($\sim$15 seconds).
\item Click ``Save'' to automatically generate and export the scene's XML configuration.
\end{enumerate}

\section{Additional Experiments}
This section extends the experiments in the main paper. The first two subsections present additional results from multi-scene closed-loop training and analyze the policy's performance and its ability to generalize to unseen scenes after training on our benchmark. The final subsection provides ablations on different observation types, examining how various observation cues impact policy performance, including: (1) \textbf{Full Observation}: includes rendered visual cones and waypoint directions in the observation space; (2) \textbf{Without Visual Cones}; (3) \textbf{Without Direction Information}.

\subsection{Task-Specific Closed-Loop Training}
\begin{figure}[!htb]
\centering
\includegraphics[width=\linewidth]{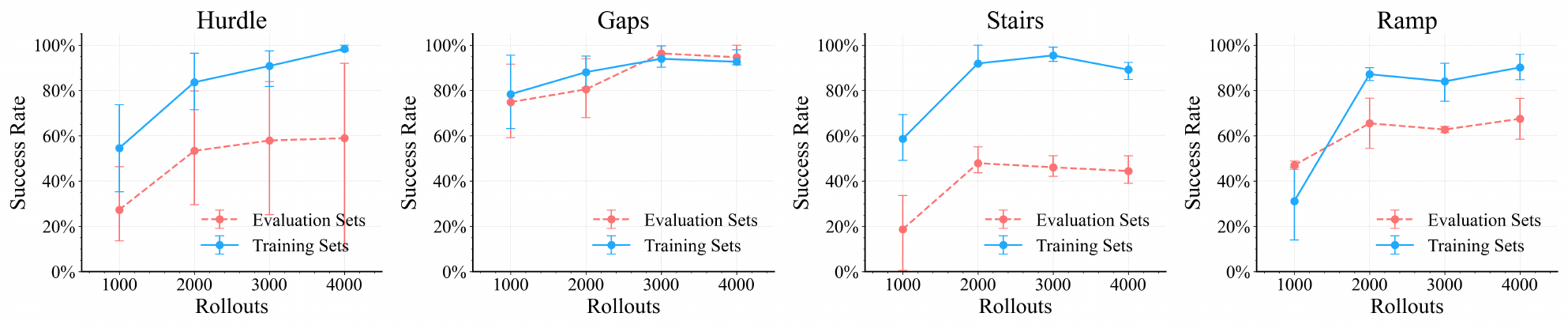}
\caption{\textbf{Task-Specific Multi-Scene Closed-Loop Training. Observation Type: RGB, Full.}}
\label{fig:supp_default_per_task}
\end{figure}

\cref{fig:supp_default_per_task} shows the results of closed-loop training for each individual task. Each policy is trained on around 10 scenes of the same task with 4 DAgger rounds. We report performance on both the training and evaluation sets; the observation type is Full Observation.
The results reveal a significant performance gap between the training and evaluation sets. This disparity indicates that closed-loop training is not effective on our benchmark, likely due to the limited number and diversity of scenes required for robust policy generalization. Nonetheless, as an evaluation benchmark, Neverwhere plays a valuable role in assessing robot policies before real-world deployment.

\subsection{Combined Task Closed-Loop Training}
\begin{figure}[!htb]
\centering
\includegraphics[width=\linewidth]{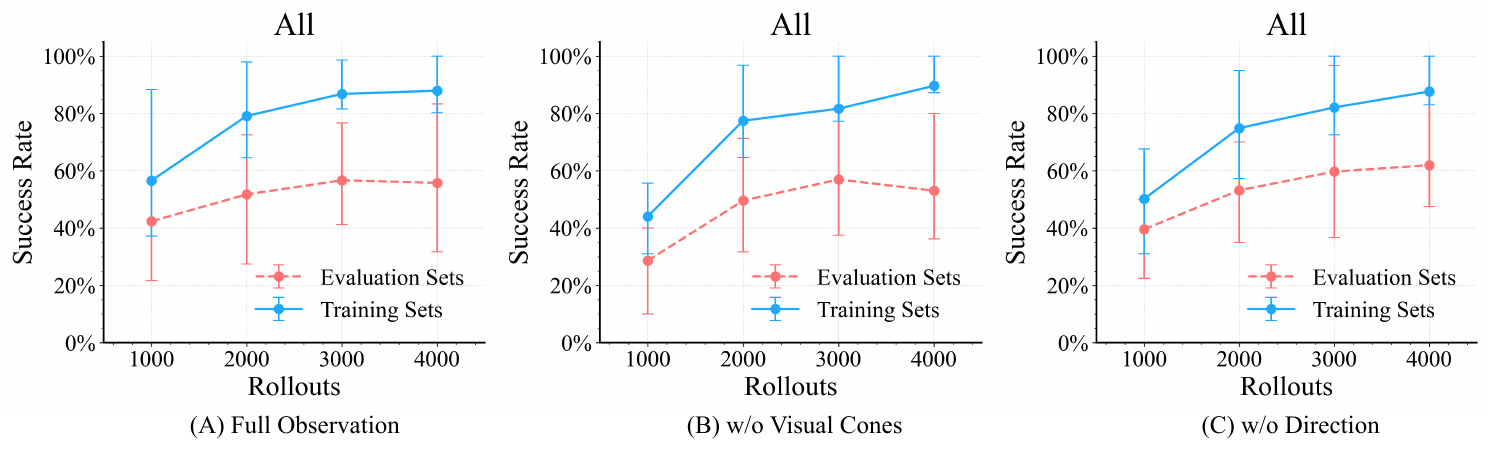}
\caption{\textbf{Combined Task Multi-Scene Closed-Loop Training. Observation Type: RGB.}}
\label{fig:supp_general_policy}
\end{figure}

We further investigate whether combining scenes across all tasks, thus increasing both scene and task diversity, can yield a more generalizable policy. We create a unified training set by merging all scenes from each task. Results are shown in \cref{fig:supp_general_policy}.
The policy achieves about 85\% success on the training set, but exhibits a 30\% performance drop on the evaluation set. This large gap suggests that, despite increased diversity, the scene set remains insufficient for training a robust visual policy. Real-world robot policy training typically requires significantly larger and more varied datasets.

\subsection{Ablation on Different Observation Cues}
\begin{figure}[!htb]
\centering
\includegraphics[width=\linewidth]{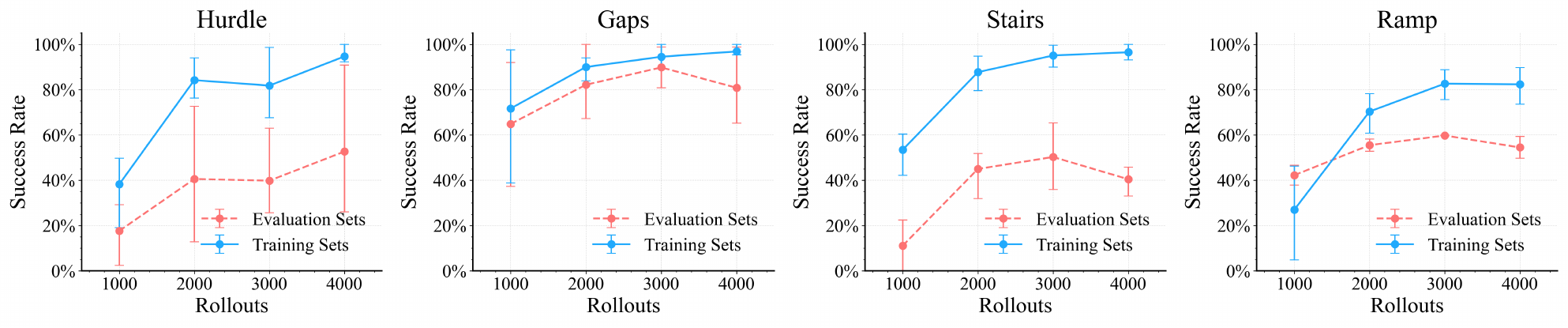}
\caption{\textbf{Task-Specific Multi-Scene Closed-Loop Training. Observation Type: RGB, w/o Visual Cones.}}
\label{fig:supp_nocones_per_task}
\end{figure}

\myparagraph{Effect of Visual Cones.}
Visual cones blended into RGB images may help the robot follow targets, but they represent artificial cues not typically available in real-world settings. To better align with real deployment conditions, we ablate this feature. Comparing performance on two tasks, Hurdle (easiest) and Ramp (hardest), under full RGB observation (\cref{fig:supp_default_per_task}) versus without visual cones (\cref{fig:supp_nocones_per_task}), we observe that policies trained \textit{with} visual cones achieve higher success rates on evaluation sets. This suggests visual cones provide a strong, scene-agnostic visual pattern that helps the policy focus on goal-relevant features and improves generalization to unseen scenes.

\begin{figure}[!htb]
\centering
\includegraphics[width=\linewidth]{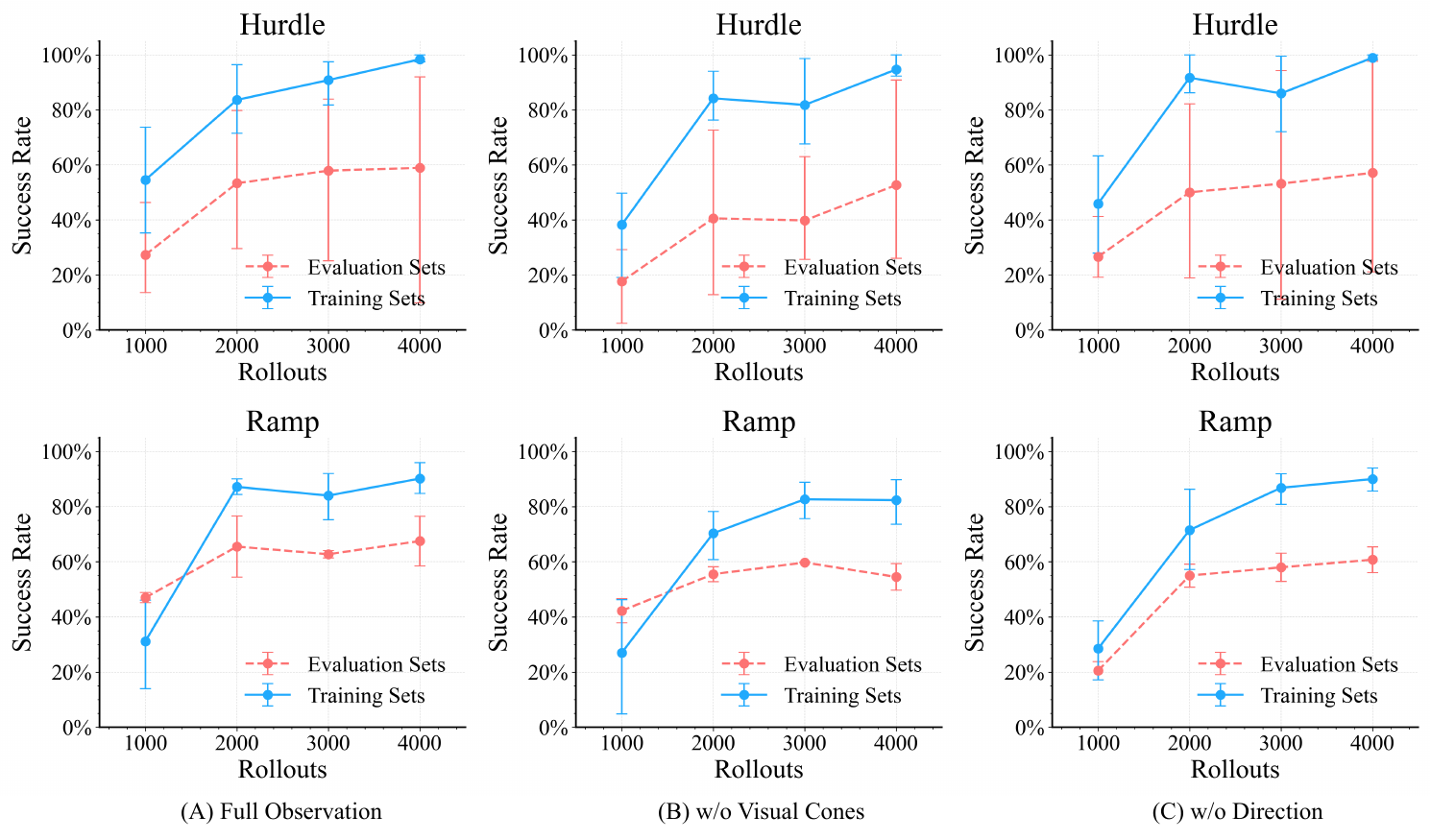}
\caption{\textbf{Ablation on Different Observation Cues.}}
\label{fig:supp_abalation_obs_hurdle_ramp}
\end{figure}

\myparagraph{Effect of Directional Information.}
We also study the impact of explicit direction input in the observation space. Comparing policies trained with direction cues (\cref{fig:supp_abalation_obs_hurdle_ramp}-(A)) and without them (\cref{fig:supp_abalation_obs_hurdle_ramp}-(C)), we find a moderate drop in evaluation performance when direction is removed. This suggests that while directional input improves performance, policies can still be trained to rely primarily on visual signals (e.g., visual cones) for navigation, albeit with some trade-off in effectiveness.

\end{document}